\documentclass[journal,twoside]{IEEEtran}
\usepackage{cite}
\usepackage{amsmath,amssymb,amsfonts}
\usepackage{algorithmic}
\usepackage{graphicx}
\usepackage{textcomp}
\usepackage{xcolor}
\usepackage{subcaption}
\usepackage{multirow}
\usepackage{hyperref}

\def\BibTeX{{\rm B\kern-.05em{\sc i\kern-.025em b}\kern-.08em
    T\kern-.1667em\lower.7ex\hbox{E}\kern-.125emX}}
\begin{document}
\title{DRBD-Mamba for Robust and Efficient Brain Tumor Segmentation with Analytical Insights}
\author{Danish Ali, Ajmal Mian, Naveed Akhtar, and Ghulam Mubashar Hassan
\thanks{
This research is supported by the Australian Government
Research Training Scholarship. The authors also gratefully acknowledge the organizers of the BraTS2023 Challenge for providing the dataset used in this research. Professor Ajmal Mian is the recipient of an ARC Future Fellowship Award (project \#FT210100268), funded by the Australian Government. Dr. Naveed Akhtar is a recipient of the ARC Discovery Early Career Researcher Award (project \#DE230101058), funded by the Australian Government. }
\thanks{Danish Ali, Ajmal Mian, and Ghulam Mubashar Hassan are with The University of Western Australia, Perth, WA 6009, Australia (e-mail: danish.ali@research.uwa.edu.au; ajmal.mian@uwa.edu.au; ghulam.hassan@uwa.edu.au).}
\thanks{Naveed Akhtar is with The University of Melbourne, Melbourne, Parkville VIC 3010, Australia (e-mail: naveed.akhtar1@unimelb.edu.au).}
\thanks{Corresponding Author: Danish Ali.}
}

\maketitle

\begin{abstract}
Accurate brain tumor segmentation is significant for clinical diagnosis and treatment but remains challenging due to tumor heterogeneity. Mamba-based State Space Models have demonstrated promising performance. However, despite their computational efficiency over other neural architectures, they incur considerable overhead for this task due to their sequential feature computation across multiple spatial axes. Moreover, their robustness across diverse BraTS data partitions remains largely unexplored, leaving a critical gap in reliable evaluation. To address this, we first propose a dual-resolution bi-directional Mamba (DRBD-Mamba), an efficient 3D segmentation model that captures multi-scale long-range dependencies with minimal computational overhead. We leverage a space-filling curve to preserve spatial locality during 3D-to-1D feature mapping, thereby reducing reliance on computationally expensive multi-axial feature scans. To enrich feature representation, we propose a gated fusion module that adaptively integrates forward and reverse contexts, along with a quantization block that improves robustness. We further propose five systematic folds on BraTS2023 for rigorous evaluation of segmentation techniques under diverse conditions and present analysis of common failure scenarios. On the 20\% test set used by recent methods, our model achieves Dice improvements of 0.10\% for whole tumor, 1.75\% for tumor core, and 0.93\% for enhancing tumor. Evaluations on the proposed systematic folds demonstrate that our model maintains competitive whole tumor accuracy while achieving clear average Dice gains of 1.16\% for tumor core and 1.68\% for enhancing tumor over existing state-of-the-art. Furthermore, our model achieves a 15x   efficiency improvement while maintaining high segmentation accuracy, highlighting its robustness and computational advantage over existing methods. The code is available at \url{https://github.com/danishali6421/DRBD_Mamba}.
\end{abstract}

\section{Introduction}
\label{sec:introduction}
\IEEEPARstart{B}{rain} tumors, particularly gliomas originating from glial cells within the central nervous system, pose a significant threat to patient survival and neurological function due to their aggressive and infiltrative nature \cite{fabian2021novel}. Precise segmentation of brain tumor is critical for accurate diagnosis, preoperative planning, radiotherapy guidance, and longitudinal monitoring. Brain tumor segmentation aims to localize anatomically distinct regions within the brain, including both healthy tissues and tumorous sub-regions such as enhancing tumor, necrotic core, and peritumoral edema. Structural and functional characteristics of these regions are derived from neuroimaging data acquired using various medical image acquisition techniques, including computed tomography (CT), magnetic resonance imaging (MRI), positron emission tomography (PET), and single-photon emission computed tomography (SPECT). Among these, magnetic resonance imaging (MRI) remains the standard for capturing brain tissue information, owing to its non-invasive nature, superior soft tissue contrast, and ability to provide multi-parametric insights critical for tumor characterization \cite{icsin2016review}. 

Multiparametric 3D MRI, comprising sequences such as T1-weighted (T1), contrast-enhanced T1-weighted (T1ce), T2-weighted (T2), and fluid-attenuated inversion recovery (FLAIR), offers diverse tissue contrasts that are essential for capturing tumor heterogeneity. During MRI acquisition, clinical 3D structural scans typically produce approximately 150 2D slices that collectively reconstruct the full brain volume \cite{chaki2022brain}. When multiple modalities, such as T1, T1ce, T2, and FLAIR are acquired, the resulting multi-parametric data becomes both volumetrically dense and diagnostically rich. However, manually inspecting each slice across all modalities to delineate healthy tissue and tumor subregions is labor-intensive, time-consuming, and subject to inter-observer variability, highlighting the need for automated and reliable segmentation methods.

Among the earliest automated brain tumor segmentation approaches, traditional machine learning methods were commonly employed to process multi-modal 3D MRI data. These techniques extracted voxel-wise intensity-based features from individual MRI sequences. Conventional classifiers, such as support vector machines and random forests, were trained on these features to generate segmentation maps \cite{bauer2011fully, lefkovits2016brain}. To improve anatomical consistency, spatial priors were incorporated using probabilistic atlases, and label fusion strategies were introduced to reduce prediction errors \cite{menze2014multimodal}. Additionally, hand-crafted spatial features derived using Gabor filter banks were integrated to capture local texture variations, while fully convolutional networks (FCNs) were employed in parallel to derive machine-learned features \cite{soltaninejad2017multimodal}. Despite these advances, traditional machine learning approaches remained heavily dependent on extensive pre-processing and hand crafted features. The quality of segmentation is closely related to the relevance and discriminative power of these handcrafted representations, often limiting their robustness in diverse imaging protocols and clinical settings \cite{khan2020cascading, qin2025btsegdiff}. These limitations ultimately motivated the transition to fully data-driven deep learning paradigms.

U-Net has emerged as a foundational deep learning architecture for medical image segmentation tasks \cite{ronneberger2015u, zhou2018unet++, dinh20231m, karthik2025unified, banerjee2025volumetric}. It introduces an encoder-decoder architecture with skip connections that effectively integrates low-level spatial detail with high-level contextual information. U-Net design has demonstrated notable success in segmenting complex anatomical patterns, particularly in the context of brain tumor subregions derived from multi-modal MRI scans. Building on its success, numerous 2D U-Net extensions, including BU-Net \cite{rehman2020bu}, Z-Net \cite{ottom2022znet}, ResU-Net \cite{metlek2023resunet+}, and Hybrid PA-NET \cite{shaheema2024explainability} have been proposed to enhance segmentation accuracy through architectural refinements. However, 2D models process mpMRI data slice by slice, often leading to inter-slice information loss and boundary artifacts once slices are stacked to form 3D segmentation maps \cite{liu2022multimodal}. In contrast, 3D models such as multi-scale 3D CNN \cite{mzoughi2020deep}, AFPNet \cite{zhou2020afpnet}, 3D FCNN \cite{anand2020brain}, TDPC-Net \cite{li2025tdpc}, and MSDMAT-BTS \cite{gao2025msdmat} operate directly on the full 3D volume, enabling more effective modeling of spatial continuity and contextual relationships across slices. Although these methods exhibit strong representational capacity, their ability to model long-range dependencies is constrained by the limited receptive fields of convolutional kernels.

The introduction of Vision Transformers (ViTs) revolutionized the field by enabling global context modeling through self-attention mechanism~\cite{dosovitskiy2020image, zhang2025bdfm}. TransBTS~\cite{wang2021transbts} pioneered transformer-based architectures for brain tumor segmentation to improve semantic understanding across spatially distant regions. However, the quadratic computational cost of standard attention mechanism poses challenges for real-world deployment \cite{ibrahim2025survey}. To address this, several efficient variants, such as window-based attention \cite{liu2021swin, hatamizadeh2021swin}, have been proposed to reduce the computational burden. However, fixed window partitioning often leads to blocking artifacts, while the shifted window strategy remains suboptimal in facilitating seamless cross-window interactions \cite{chen2023activating}. 

Recent research has increasingly explored more scalable alternatives. Among them, State Space Models (SSMs), particularly Mamba, have shown strong potential, offering linear time complexity and effective long-range context modeling across spatial sequences \cite{zhu2024vision}. However, 3D sequence modeling is challenging: naive row-major flattening breaks 3D spatial locality. To mitigate this, SegMamba \cite{xing2024segmamba} applies tri-orientation Mamba (ToM) along three anatomical directions, with ToM blocks integrated at every encoder stage to capture multi-scale sequential features. While tri-axial feature extraction and fusion enriches the representations, it incurs substantial computational and memory overhead, limiting applicability in resource-constrained clinical settings. 

Beyond architectural considerations of models, another limitation lies in the evaluation protocol for the task of brain tumor segmentation. Many prior studies adopt ad hoc {\em random} (training/validation/testing) splits \cite{baid2021rsna} such as SegMamba \cite{xing2024segmamba} (70/10/20), DB-Trans \cite{zeng2023dbtrans} (67/16/17), VcaNet \cite{pan2025vcanet} (80/10/10), and SDV-TUNet \cite{zhu2024sparse} (80/15/5), which limits comparability and can misrepresent model robustness. Moreover, these studies report results on a single split (no k-fold cross-validation) and rarely disclose the test-set composition (e.g., ET/NC/ED volume histograms), hindering reproducibility and leaving comparisons vulnerable to split-induced bias. Compounding this issue, the inherent distributional variability within the BraTS dataset \cite{baid2021rsna} makes random splits less representative of true model generalization. Consequently, current evaluation protocols limit meaningful insights into how different models perform under diverse data distributions.

To address the above limitations, we propose dual-resolution bi-directional Mamba, an efficient segmentation architecture designed to capture long-range dependencies at multiple scales while maintaining low computational cost. Furthermore, we employ a robust and reliable evaluation protocol based on stratified k-fold cross-validation \cite{michelucci2024model} to ensure fair and generalizable performance assessment. The main contributions of this paper are as follows:
\begin{itemize}
    \item We propose a novel dual-resolution Mamba, where Mamba blocks are selectively placed at two key network locations only: one in the bottleneck and another in the skip connection from the preceding encoder stage to capture multi-scale global context with minimal computational burden. 
    \item We employ a Z-order (Morton) space-filling curve to map 3D features into 1D sequences while preserving spatial locality and avoiding the dyadic padding overhead of Hilbert curves. On top of this representation, we propose bidirectional Mamba to perform forward and reverse scans, and propose a channel-wise gating mechanism to adaptively fuse the two streams, enabling discriminative modeling of long-range dependencies.
    \item We propose a vector quantization module that discretizes
    Mamba-encoded feature representations, thereby improving model robustness to noise by regularizing the latent representation.
    \item We propose systematic five folds in which data is organized based on average foreground (tumorous region) intensity variation, with each fold exhibiting different distributions of tumor volume, providing a fair and reproducible evaluation of model robustness under clinically diverse conditions.
\end{itemize}
Extensive evaluation on the BraTS 2023 dataset \cite{baid2021rsna}, demonstrates that our method improves segmentation accuracy compared to state-of-the-art models \cite{hatamizadeh2021swin, he2023swinunetr, xing2024segmamba, xing2025segmamba}, with reduced computational cost. Furthermore, cross-fold validation demonstrates our model’s promising performance across varied data distributions, reflecting robustness to variations in tumor subregion intensity and volume. 
\section{Related Work}\label{relwork}
This section presents an overview of the methodologies commonly used in brain tumor segmentation. We first review CNN-based approaches. We then summarize hybrid architectures that combine convolutional encoder-decoder backbones with either vanilla or window-based attention mechanisms to capture long-range dependencies, as well as emerging Mamba-based designs that offer a promising alternative for efficient sequence modeling.
\subsection{CNN-based Methods}\label{cnn}
CNNs have been a robust baseline for brain tumor segmentation, owing to their ability to extract rich, hierarchical features across multiple spatial scales \cite{zhang2025annular}. Building on this foundation, several 3D CNN variants have been proposed to better handle the heterogeneity of tumor subregions. For instance, the multi-branch attention network (MBANet) \cite{cao2023mbanet} integrates channel and spatial attention into skip connections, enhancing volumetric feature fusion while reducing computational complexity through group convolutions. The hierarchical multi-scale network (HMNet) \cite{zhang2023hmnet} captures tumor structures at multiple resolutions, improving adaptability to brain tumors with diverse morphologies and spatial characteristics. Extending this multi-scale strategy, the multi-scale residual U-Net (mResU-Net) \cite{li2024mresu} integrates dilated convolutions with multiple scales in both the encoder and decoder, expanding the receptive field to capture features across diverse spatial scales and improving segmentation accuracy for targets with varied anatomical extents.
\subsection{Transformer-based Hybrid Architectures}\label{trans}
CNN-based methods, even with multi-scale and dilated convolutions, can expand receptive fields but still struggle to model explicit long-range dependencies. To address this, recent methods adopt a hybrid design that integrates Transformer-based attention mechanisms with CNN models, combining the local feature extraction strength of CNNs with the global context modeling capability of Transformers. TransBTS \cite{wang2021transbts}, marked the pioneering attempt to incorporate Transformer bottleneck into 3D CNN for the segmentation of brain tumors in multi-modal MRI. In UNETR \cite{hatamizadeh2022unetr}, a Transformer block was incorporated into every encoder layer to capture multi-scale global information across multiple spatial resolutions. However, this design incurs substantial computational cost, as attention operations must be performed at each resolution. To address this challenge, Jia et al. proposed an enhanced variant of TransBTS called BiTr-Unet \cite{jia2021bitr}. This network embeds transformer blocks in the final two encoder layers to extract global information at two different scales, effectively mitigating the limitations of the bottleneck-only design while achieving superior computational efficiency compared to UNETR. Despite these efficiency gains, each Transformer block still computes attention over the entire sequence, which remains the bottleneck for large volumes.

To address this, window-based attention mechanisms have been introduced, where 3D features are partitioned into fixed-size windows and attention is computed locally within each window \cite{hatamizadeh2021swin}. To enhance cross-window interaction, Zeng et al. \cite{zeng2023dbtrans} incorporated a shuffle window cross-attention module that explicitly relates each window to spatially distant, non-overlapping windows.  Despite their efficiency, these architectures typically perform window partitioning and subsequent patch merging at every encoder stage after attention computation, which introduces blocking artifacts that degrade segmentation performance \cite{chen2023activating}. 
\subsection{Mamba-based Methods}\label{mamba_models}
 Transformer-based hybrid models \cite{wang2021transbts, zeng2023dbtrans} effectively model global dependencies, but their quadratic computational complexity limits their scalability \cite{ibrahim2025survey}. To overcome this, 
 Seg-Mamba \cite{xing2024segmamba} and Seg-Mamba V2 \cite{xing2025segmamba} adopt state-space models (SSMs) to capture long-range dependencies with linear complexity. Unlike Transformers \cite{vaswani2017attention}, where each token attends to all others, SSMs process tokens sequentially. Extending this to 3D volumes is nontrivial: naive 1D flattening disrupts spatial locality \cite{xie2024quadmamba}. 

 To address this, SegMamba \cite{xing2024segmamba} forms sequences along three directions (forward, reverse, and inter-slice), applies Mamba in each direction, and fuses features at every encoder stage. SegMamba-V2 \cite{xing2025segmamba} further extends this design by introducing an improved Tri-Oriented Ortho Mamba (ToOM) module that jointly considers feature interactions from multiple directions and anatomical planes (axial, coronal, and sagittal) to achieve a better understanding of 3D medical images. Although more efficient than full attention, Mamba at early high-resolution stages remains computationally expensive, and processing multiple orientations compounds computational complexity. Moreover, both SegMamba \cite{xing2024segmamba} and SegMamba-V2 \cite{xing2025segmamba} rely on row-major flattening, which weakens 3D spatial locality as depth-adjacent voxels become distant in the sequence, leading to reduced sequence modeling efficacy.

In contrast to the above, our  solution achieves efficient global context modeling while eliminating reliance on multi-axial Mamba by adopting a space-filling curve (SFC) ordering that preserves 3D spatial locality and improves sequence representation. Furthermore, we propose five systematic folds and perform cross-fold validation to evaluate robustness across diverse data distributions.
\section{The Proposed Methodology}
\begin{figure*}[t]
\centerline{\includegraphics[width=\textwidth]{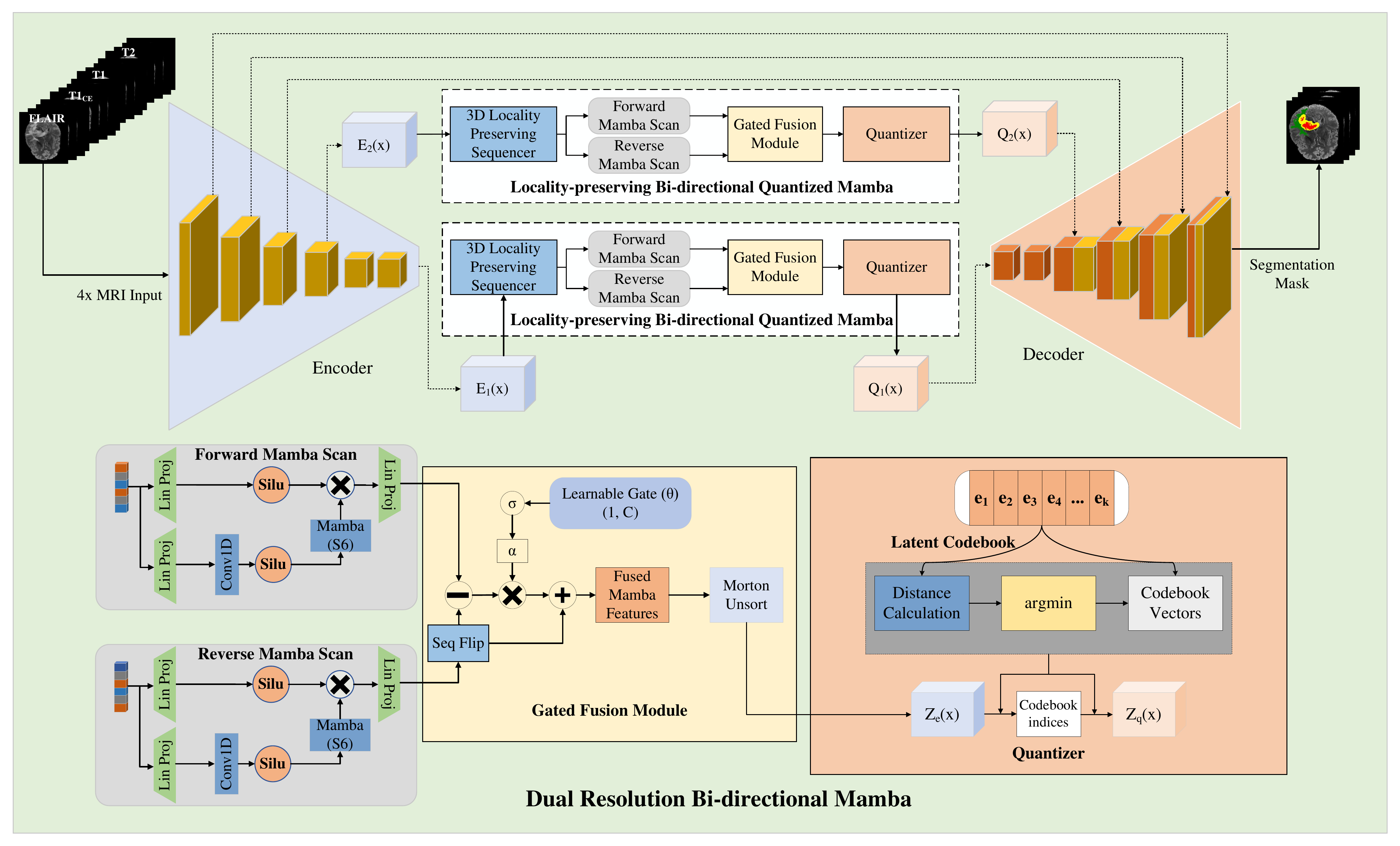}}
\caption{The overall architecture of the proposed dual-resolution bi-directional Mamba (DRBD-Mamba) with Mamba blocks placed in the bottleneck and skip connection to capture multi-scale context.}
\label{bi-mamba}
\end{figure*}

\subsection{Overview}
Fig. \ref{bi-mamba} illustrates the proposed dual-resolution bi-directional Mamba (DRBD-Mamba) which follows an encoder–decoder design. The network takes a multi-modal MRI volume of shape $x \in \mathbb{R}^{4 \times H \times W \times D}$ as an input, where the four modalities (T1, T1ce, T2, and FLAIR) are stacked along the channel dimension.  

The input is passed through a CNN encoder with six 3D convolution stages that progressively down-sample the feature maps and extract hierarchical local representations. The output of the $i$-th encoder stage is defined as $e_i \in \mathbb{R}^{C_i \times H_i \times W_i \times D_i}$, where $i \in [1,6]$ and $C_i, H_i, W_i, D_i$ denote the number of channels and spatial dimensions at stage $i$. The core component of our architecture is the locality-preserving bi-directional quantized Mamba module, designed to capture long-range dependencies within high-dimensional semantic features. Specifically, one bi-directional Mamba module is incorporated into the bottleneck, and another is integrated into the skip connection from the preceding encoder stage. This design choice is motivated by the complementary strengths of convolution and sequence modeling: convolution excels at aggregating fine-grained local features in higher-resolution stages, whereas Mamba effectively models long-range dependencies in lower-resolution feature maps with richer semantic details, all while maintaining linear computational complexity. The features from the forward and reverse Mamba scans are adaptively fused to derive rich semantic representations.

To further improve robustness, we introduce a quantization block that discretizes the features extracted by the Mamba module. By constraining the feature space to a fixed number of discrete embeddings, this block reduces sensitivity to noise. Finally, the decoder, composed of convolution blocks with upsampling, reconstructs the high-dimensional features into the final segmentation output.

\subsection{Mamba: Input-Dependent State Space Models}\label{s6}
State Space Models (SSMs) provide a principled formulation for sequence modeling, where inputs evolve through an internal state-transition process. In particular, Structured State Space Models (S4) introduce a parameterization of continuous-time linear time-invariant (LTI) systems that makes them highly expressive while maintaining computational efficiency for large-scale deep learning. However, S4 relies on fixed transition parameters ($A \in \mathbb{R}^{N \times N}$ (state transition), $B \in \mathbb{R}^{N \times 1}$ (input mapping), $C \in \mathbb{R}^{1 \times N}$ (state projection), and the discretization step size $\Delta > 0$) that remain constant across the sequence. 

In our work, we employ Mamba-based State Space Models (S6), which extend the idea of S4 by introducing input-dependent parameters that allow the state evolution to adapt dynamically to the input context. In particular, the matrices $B$, $C$, and the step size $\Delta$ are parameterized as functions of the input $x_k$, i.e., $B(x_k)$, $C(x_k)$, and $\Delta(x_k)$. This input-dependent formulation enables context-aware modulation of the system. The resulting recurrence relation becomes:
\begin{equation}
    h_k = \bar{A} h_{k-1} + \bar{B}(x_k) \, x_k, \quad y_k = C(x_k) h_k,
\end{equation}
where $\bar{A} \in \mathbb{R}^{N \times N}$ is the discretized state transition matrix, $\bar{B}(x_k) \in \mathbb{R}^{N \times 1}$ maps the input $x_k$ into the hidden state, and $C(x_k) \in \mathbb{R}^{1 \times N}$ projects the hidden state back to the output $y_k \in \mathbb{R}$. 
Both matrices $\bar{B}(x_k)$ and $C(x_k)$ are functions of the input. This design can be interpreted as a form of selective SSM, where the model determines at each time step, how strongly the new input influences the hidden state. In practice, Mamba employs efficient parallelization and linear-time operations, preserving the scalability of S4 while substantially improving its expressiveness. By combining structured recurrence with input-adaptive modulation, Mamba bridges the gap between classical SSMs and modern sequence architectures, providing a powerful alternative to attention mechanism.
\subsection{Network Encoder}  
The network encoder (Fig. \ref{bi-mamba}) is designed to hierarchically extract both local and global features from multi-modal MRI inputs. The input volume of shape $(4, H, W, D)$, where four modalities are stacked along the channel dimension, is first processed by a convolution backbone composed of six stages. These stages progressively capture local features across multiple resolutions through down-sampling, yielding feature maps of sizes: $(16, H, W, D)$, $(32, \tfrac{H}{2}, \tfrac{W}{2}, \tfrac{D}{2})$, $(64, \tfrac{H}{4}, \tfrac{W}{4}, \tfrac{D}{4})$, $(128, \tfrac{H}{8}, \tfrac{W}{8}, \tfrac{D}{8})$, and $(256, \tfrac{H}{16}, \tfrac{W}{16}, \tfrac{D}{16})$.  

At the sixth stage, the low-resolution features are projected into higher-dimensional representations using an additional convolution layer while maintaining the same spatial resolution, resulting in feature maps of $(512, \tfrac{H}{16}, \tfrac{W}{16}, \tfrac{D}{16})$. These semantically richer feature representations benefit from effective modeling of long-range dependencies to capture contextual information. To this end, we employ a locality-preserving bi-directional quantized Mamba block in the bottleneck, which effectively models long-range sequence dependencies while preserving spatial coherence. This design allows the encoder to integrate both fine-grained local information and global context in a computationally efficient manner.
\subsection{Locality-Preserving Bi-Directional Quantized Mamba}
The 3D feature embedding obtained from the final convolution stage of the encoder is processed by the locality-preserving bi-directional quantized Mamba block which comprises of four main components: 3D sequencer, bi-directional Mamba, gated fusion module and vector quantizer. The architectural detail of each component is explained below.
\subsubsection{3D Locality Preserving
Sequencer}\label{3dlps}
The effectiveness of sequence models such as Mamba for 3D medical data depends on mapping volumetric features into a sequential form that retains spatial locality. A naive approach is to flatten along each axis and compute Mamba features independently before fusing them, but this introduces substantial computational overhead \cite{hamilton2007compact, bohm2020space}. Space-filling curves provide an alternative by mapping multidimensional data into a 1D sequence while better preserving spatial neighborhoods.

Among such methods, standard Hilbert curves offer strong theoretical guarantees of spatial locality. However, their recursive construction is only defined for grids where each dimension follows a binary subdivision (e.g., 2, 4, 8, ...). As a result, feature maps with arbitrary spatial resolutions, such as those produced by our encoder where the latent spatial resolution (e.g., $\tfrac{H}{16} \times \tfrac{W}{16} \times \tfrac{D}{16}$) is not powers of two, must be padded to the next valid grid size. This padding increases memory usage and introduces substantial computational overhead.

We therefore use Morton (Z-order) mapping in our network because it achieves locality preservation through bit interleaving, supports arbitrary grid sizes without padding, and remains lightweight to compute. The Morton index for a voxel at coordinates $(x,y,z)$ is computed by interleaving the bits of the coordinates:
\begin{multline}\label{mor}
\text{Morton}(x,y,z) = \sum_{i=0}^{b-1} 
 \big( (x_i \ll (3i)) + (y_i \ll (3i+1)) \\
 + (z_i \ll (3i+2)) \big),
\end{multline}
where $x_i, y_i, z_i$ denote the $i$-th bits of the coordinates, and $\ll$ is the bit-shift operator. The encoder feature maps $E(x) \in \mathbb{R}^{B \times 512 \times \tfrac{H}{16} \times \tfrac{W}{16} \times \tfrac{D}{16}}$
are flattened and permuted following the Morton sequence defined in \eqref{mor}, yielding
\begin{equation}
s = \text{Morton}(E(x)) \in \mathbb{R}^{B \times L \times 512},
\qquad L = \tfrac{H}{16} \cdot \tfrac{W}{16} \cdot \tfrac{D}{16},
\end{equation}
where $s$ denotes the Morton-ordered flattened sequence. This representation ensures that voxels close in 3D space remain adjacent in the 1D sequence, thereby preserving spatial locality while avoiding the computational overhead of axis-wise flattening.
\subsubsection{Bi-directional Mamba}
The Morton-ordered flattened sequence $s$ obtained from the 3D sequencer block is fed into a bi-directional Mamba block, which models long-range dependencies in both forward and reverse directions while maintaining linear computational complexity. In the forward direction, Mamba (see Sec. \ref{s6}) processes this sequence (s) as a state-space recurrence over discrete steps $k = 1,\dots,L$:
\begin{equation}
    h^{\text{fwd}}_k = \bar{A} h^{\text{fwd}}_{k-1} + \bar{B}(s_k)\, s_k, 
    \qquad 
    y^{\text{fwd}}_k = C(s_k)\, h^{\text{fwd}}_k .
\end{equation}

For the reverse pass, the sequence is flipped along the token dimension, processed through the same recurrence, and flipped back to align with the forward ordering:
\begin{equation}
    y_k^{\text{rev}} = \text{Flip}\big(\text{Mamba}(\text{Flip}(s_k))\big).
\end{equation}

This bi-directional formulation enables each token representation to aggregate information from both temporal directions. Specifically, the forward Mamba pass captures contextual dependencies from preceding tokens, whereas the reverse pass extracts complementary information from subsequent tokens. Consequently, the bi-directional Mamba integrates both past and future context, facilitating more comprehensive modeling of long-range dependencies across the volumetric sequence.
\subsubsection{Gated Fusion Module}
The bi-directional Mamba captures contextual information from both temporal directions independently. The proposed gated fusion module is designed to fuse the forward ($y^{\text{fwd}}$) and reverse ($y^{\text{rev}}$) representations on a per-channel basis through a learnable gating mechanism. A parameter vector $\theta \in \mathbb{R}^{512}$ is optimized during training, and the fusion weights are defined as $\alpha = \sigma(\theta)$. This gating mechanism enables the model to adaptively decide, for each channel, whether the forward or reverse context should dominate. Instead of uniformly averaging both passes, the gate selectively emphasizes the more informative direction at every feature dimension, leading to more discriminative and semantically meaningful representations. The fused Mamba features are computed as:
\begin{equation}
    y_k = \alpha \odot y^{\text{fwd}}_k \;+\; (1-\alpha) \odot y^{\text{rev}}_k,
\end{equation}
where $\odot$ denotes element-wise multiplication, $y_k$ is the fused feature representation at sequence index $k$, $y^{\text{fwd}}_k, y^{\text{rev}}_k$ are the forward and reverse scan features obtained from the bidirectional Mamba, $\alpha$ is the gating vector that adaptively controls the contribution of forward and reverse features per channel dimension. Finally, the fused sequence $y \in \mathbb{R}^{B \times L \times 512}$ is mapped back to the spatial domain using the inverse Morton permutation, resulting in
\begin{equation}
    Y = \text{Morton}^{-1}(y) \in \mathbb{R}^{B \times 512 \times \tfrac{H}{16} \times \tfrac{W}{16} \times \tfrac{D}{16}} .
\end{equation}

This design not only integrates bidirectional contextual information into a unified spatial representation, but also offers a lightweight and computationally efficient alternative to tri-axial scanning. Consequently, the resulting feature map preserves structural fidelity while enriching the semantic content, ultimately facilitating more accurate downstream predictions. 
\subsubsection{Quantizer}
The fused representations provide rich contextual encoding, which is further structured in the latent space through a discretization mechanism. Specifically, we introduce a vector quantizer (VQ) that transforms the continuous Mamba features ($Y$) into a finite set of embedding vectors. This transformation from continuous to discrete space improves robustness to noise and encourages the learning of more generalizable feature representations.

The vector quantizer is parameterized by a codebook $E = { e_k }_{k=1}^K$, comprising $K$ embedding vectors of dimensionality $D$. Given an encoded latent representation $Y(x)$, quantization is performed by assigning it to the closest entry in the codebook based on Euclidean distance:

\begin{equation}
Q(x) = \underset{e_k \in E}{\arg\min} \left\| Y(x) - e_k \right\|_2 ,
\end{equation}
where $Y(x)$ and $Q(x)$ denote the latent representations before and after quantization, respectively, and $e_k \in \mathbb{R}^D$ corresponds to the $k_{th}$ codebook vector. We employ a vector quantization module at the bottleneck using a codebook with $K=512$ embedding vectors of dimensionality $D=512$. The decoder reconstructs the segmentation mask from these quantized latent features through hierarchical upsampling, progressively recovering spatial resolution while preserving semantic integrity. 

Beyond conventional encoder-decoder skip connections, which predominantly transfer shallow local details, we embed a locality-preserving bidirectional quantized Mamba module within the skip connection of the preceding encoder stage at feature resolution $\bigl(\tfrac{H}{8}, \tfrac{W}{8}, \tfrac{D}{8}\bigr)$. The quantization module at this stage employs a codebook with $K=256$ embedding vectors of dimensionality $D=128$. This design enriches the contextual information propagated from this encoder stage to the decoder by aggregating sequential information within the skip
connection, while simultaneously reducing sensitivity to noise in the mid-level feature representations. However, as sequence length grows substantially at higher-resolution encoder stages, applying Mamba to these skip connections becomes expensive despite its linear complexity. We therefore confine its use to the bottleneck and deepest skip connection, resulting in a dual-resolution design that captures global semantics while maintaining efficiency.

The proposed model is trained using the Cross Entropy Dice loss $\mathcal{L}_{\text{CEDice}}$, while a straight-through estimator is applied to allow gradients to propagate through the non-differentiable quantization step. The codebook vectors are updated using the exponential moving average (EMA), ensuring stability. Additionally, a commitment loss is introduced to encourage the sequence features (Y) produced by the Mamba block to remain close to the chosen codebook embeddings:
\begin{equation}
\mathcal{L}_{\text{ct}} = \left\| Y(x) - s_g[Q(x)] \right\|_2^2,
\end{equation}
where $s_g$ denotes the stop-gradient operator. The overall training objective combines the Cross Entropy Dice loss, quantization loss, and commitment loss, enabling the VQ-enhanced Mamba features to provide discrete, noise-resilient latent embeddings that are semantically expressive and lead to accurate segmentation.
\section{Experiments}\label{ressec}
This section presents the experimental setup, implementation details, and comprehensive results, including performance comparison with recent state-of-the-art methods.
\begin{table}[]
\centering
\setlength{\tabcolsep}{2pt}
\caption{Quantitative comparison of the proposed model with state-of-the-art methods on the BraTS2023 dataset. Dice similarity coefficient (Dice \%) and 95th percentile Hausdorff distance (HD95 mm) are reported for whole tumor (WT), tumor core (TC), and enhancing tumor (ET). Higher Dice and lower HD95 indicate better performance. The results in bold represent the best performances.}
\label{tab:brats2023_results}
\begin{tabular}{lcccccc}
\hline
\multirow{2}{*}{Methods} & \multicolumn{2}{c}{WT} & \multicolumn{2}{c}{TC} & \multicolumn{2}{c}{ET} \\
\cline{2-3} \cline{4-5} \cline{6-7}
 & Dice $\uparrow$ & HD95 $\downarrow$ & Dice $\uparrow$ & HD95 $\downarrow$ & Dice $\uparrow$ & HD95 $\downarrow$ \\
\hline
SegResNet \cite{myronenko20183d}     & 92.02 & 4.07 & 89.10 & 4.08 & 83.66 & 3.88 \\
UX-Net \cite{lee20223d}        & 93.13 & 4.56 & 90.03 & 5.68 & 85.91 & 4.19 \\
MedNeXt \cite{roy2023mednext}       & 92.41 & 4.98 & 87.75 & 4.67 & 83.96 & 4.51 \\
\hline
UNETR \cite{hatamizadeh2022unetr}         & 92.19 & 6.17 & 86.39 & 5.29 & 84.48 & 5.03 \\
Swin-UNETR \cite{hatamizadeh2021swin}    & 92.71 & 5.22 & 87.79 & 4.42 & 84.21 & 4.48 \\
Swin-UNETR-V2 \cite{he2023swinunetr} & 93.35 & 5.01 & 89.65 & 4.41 & 85.17 & 4.41 \\
nnFormer \cite{zhou2021nnformer}      & 91.15 & 5.65 & 85.94 & 5.31 & 78.73 & 5.09 \\
\hline
SegMamba \cite{xing2024segmamba}      & 93.03 & 4.17 & 90.26 & \textbf{3.87} & 86.53 & 4.30 \\
SegMamba-V2 \cite{xing2025segmamba}   & 93.15 & \textbf{3.71} & 90.16 & 4.02 & 86.64 & \textbf{3.56} \\
Proposed   & \textbf{93.45} & 5.41 & \textbf{92.01} & 4.52 & \textbf{87.97} & 4.89 \\
\hline
\end{tabular}
\end{table}
\subsection{Dataset}
We evaluate the proposed model on the BraTS 2023 benchmark dataset, which consists of 1251 multi-institutional 3D brain MRI cases. Each case contains four modalities: T1-weighted (T1), contrast-enhanced T1-weighted (T1ce), T2-weighted (T2), and fluid-attenuated inversion recovery (FLAIR). Following recent literature \cite{myronenko20183d, lee20223d, hatamizadeh2021swin, xing2024segmamba, xing2025segmamba}, we adopt a 70/10/20 split for training, validation, and testing, keeping the test set consistent with prior studies for fair comparison. The ground truth segmentation masks include four labels: background, peritumoral edema (ED), necrotic and non-enhancing tumor core (NCR), and enhancing tumor (ET). In accordance with the BraTS protocol, results are reported on three composite regions: whole tumor (WT = ED+NCR+ET), tumor core (TC = NCR+ET), and enhancing tumor (ET).
\begin{table*}[t]
\centering
\caption{Performance comparison between SwinUNETR \cite{hatamizadeh2021swin} and the proposed DRBD-Mamba on the BraTS2023 dataset using the same random five-fold splits as defined by SwinUNETR \cite{hatamizadeh2021swin}. The proposed DRBD-Mamba consistently outperforms SwinUNETR across all folds. Best results with the highest Dice scores (\%) and lowest HD95 (mm) are highlighted in \textbf{bold}.}
\label{tab:random_folds}
\setlength{\tabcolsep}{6pt}
\begin{tabular}{l|cccccc|cccccc}
\hline
\multirow{2}{*}{Random Folds} & \multicolumn{6}{c|}{\textbf{SwinUNETR}} & \multicolumn{6}{c}{\textbf{Proposed DRBD-Mamba}} \\
\cline{2-13} 
& \multicolumn{2}{c}{WT} & \multicolumn{2}{c}{TC} & \multicolumn{2}{c|}{ET} 
& \multicolumn{2}{c}{WT} & \multicolumn{2}{c}{TC} & \multicolumn{2}{c}{ET} \\
\cline{2-13}
& Dice & HD95 & Dice & HD95 & Dice & HD95 & Dice & HD95 & Dice & HD95 & Dice & HD95 \\
\hline
Fold 1 & 91.56 & 8.52 & 88.31 & 5.31 & \textbf{84.13} & 4.32 
       & \textbf{92.29} & \textbf{6.85} & \textbf{89.19} & \textbf{4.70} & 84.10 & \textbf{3.90} \\
Fold 2 & 92.73 & 7.43 & 90.29 & 4.60 & 87.49 & 3.81 
       & \textbf{93.27} & \textbf{6.13} & \textbf{90.58} & \textbf{4.04} & \textbf{87.90} & \textbf{3.24} \\
Fold 3 & 92.41 & 6.89 & \textbf{90.21} & 5.12 & 87.19 & \textbf{4.36}
       & \textbf{92.51} & \textbf{5.38} & 89.68 & \textbf{5.09} & \textbf{87.37} & 4.63 \\
Fold 4 & 91.78 & 8.14 & 89.39 & 5.16 & 85.35 & 3.95 
       & \textbf{92.41} & \textbf{6.45} & \textbf{90.62} & \textbf{5.06} & \textbf{86.00} & \textbf{3.66} \\
Fold 5 & 92.18 & 6.18 & 89.55 & \textbf{4.76} & 83.74 & \textbf{3.71} 
       & \textbf{92.66} & \textbf{4.96} & \textbf{89.64} & 4.83 & \textbf{85.66} & 3.82 \\
\hline
Mean   & 92.13 & 7.43 & 89.55 & 4.99 & 85.58 & 4.03 
       & \textbf{92.63} & \textbf{5.95} & \textbf{89.94} & \textbf{4.74} & \textbf{86.21} & \textbf{3.85} \\
Std    & 0.47  & 0.94 & 0.80  & 0.30 & 1.72  & 0.30 
       & 0.38  & 0.77 & 0.63  & 0.43 & 1.50  & 0.50 \\
\hline
\end{tabular}
\end{table*}
\subsection{Evaluation Metrics}
Performance is quantitatively assessed using two widely adopted metrics: Dice similarity coefficient (Dice) and 95th percentile Hausdorff distance (HD95). Dice measures volumetric overlap between predicted and reference segmentations, while HD95 evaluates boundary accuracy.
\subsection{Implementation Details}
The proposed model is implemented in PyTorch and trained for 500 epochs on a single NVIDIA RTX 3090 GPU (24 GB) with a batch size of 3. A crop size of $160\times160\times144$ voxels is used during training. Inference is performed using sliding window evaluation with 50\% overlap. Optimization is carried out using Adam Optimizer with an initial learning rate of $1\times10^{-4}$ and weight decay of $1\times10^{-4}$. To improve robustness and generalization, data augmentation is done including random flipping along three axes (probability 0.5), random rotation (0.5), intensity shifting (0.1), and intensity scaling (0.1).
\subsection{Performance Comparison}
We compare the proposed model against recent state-of-the-art approaches on the BraTS 2023 dataset, ensuring identical test set for fairness. The state-of-the-art models include CNN-based models including SegResNet \cite{myronenko20183d}, UX-Net \cite{lee20223d}, and MedNeXt \cite{roy2023mednext}; transformer-based models including UNETR \cite{hatamizadeh2022unetr}, Swin-UNETR \cite{hatamizadeh2021swin}, Swin-UNETR-V2 \cite{he2023swinunetr}, and nnFormer \cite{zhou2021nnformer}; and Mamba-based models including SegMamba \cite{xing2024segmamba} and SegMamba-V2 \cite{xing2025segmamba}. 
The results for SegMamba \cite{xing2024segmamba} and SegMamba-V2 \cite{xing2025segmamba} are reproduced as these two are competitive baselines emphasizing computational efficiency under an identical 500-epoch training setup as our model, while results for the other methods are taken from the literature. 

The quantitative results are presented in Table \ref{tab:brats2023_results}. It can be observed that the proposed model performs better than the compared state-of-the-art methods and achieves consistent improvements across tumor subregions, with Dice gains of 0.10\% for WT, 1.75\% for TC, and 1.33\% for ET. For HD95, the performance is comparable to the best results, indicating competitive boundary precision. The Dice score improvements are particularly pronounced for the more challenging TC and ET subregions.

\subsection{Cross-Fold Validation and Comprehensive Analysis}
Most prior works \cite{hatamizadeh2022unetr, myronenko20183d, xing2024segmamba, xing2025segmamba} report results on a single random 20\% test set, which may not fully capture robustness under heterogeneous distributions. To ensure a more reliable assessment \cite{michelucci2024model}, we perform five-fold cross-validation on BraTS2023 using the random folds defined by Swin-UNETR \cite{hatamizadeh2021swin}. To the best of our knowledge, Swin-UNETR \cite{hatamizadeh2021swin} is the only study in the literature that has performed cross-fold validation on the BraTS dataset and publicly shared its folds. For a fair comparison, both our model and Swin-UNETR are evaluated after 300 training epochs on the same folds. From Table \ref{tab:random_folds}, it can be clearly observed that our method consistently outperforms Swin-UNETR across all the random folds, and provides new state-of-the-art results. 

With thorough analysis of the obtained results, we observe that the random folds of \cite{hatamizadeh2021swin} show limited variability, with average standard deviations of 0.38/0.63/1.50 (Dice) and 0.77/0.43/0.50 (HD95) for WT/TC/ET respectively. We also observe that several common low-performing cases across all folds for both our proposed model and Swin-UNETR are the same, which correspond to smaller tumor volumes. In addition, as the BraTS dataset is collected from multiple institutions and scanners, the foreground (tumor-region) intensity variation differs considerably across the cases \cite{wrobel2020intensity}. 
These observations motivate us to devise a more systematic method for designing the folds since the existing random folds 
do not adequately account for scanner-dependent intensity variation or differences in tumor volume across cases. Therefore, to ensure a consistent and unbiased evaluation of segmentation performance across the dataset, we propose systematically designed well-balanced folds. This design aligns with Michelucci’s perspective \cite{michelucci2024model}, which emphasizes that stratified folds enable fairer performance estimation and more reliable understanding of model generalizability in heterogeneous datasets. 
\begin{table*}[t]
\centering
\caption{Systematic five-fold cross-validation results on BraTS2023, comparing the proposed model against state-of-the-art baselines (SegMamba \cite{xing2024segmamba} and SegMambaV2 \cite{xing2025segmamba}). Dice score (\%) is reported for whole tumor (WT), tumor core (TC), and enhancing tumor (ET). Best results per fold are highlighted in bold. Numbers in green/red denote performance gains/decrements of the proposed model compared to the best baseline.}
\label{tab:segmamba_comparison}
\setlength{\tabcolsep}{6pt}
\begin{tabular}{l|ccc|ccc|ccc}
\hline
\multirow{2}{*}{Systematic Folds} 
& \multicolumn{3}{c|}{SegMamba \cite{xing2024segmamba}} 
& \multicolumn{3}{c|}{SegMambaV2 \cite{xing2025segmamba}} 
& \multicolumn{3}{c}{Proposed} \\
\cline{2-10}
& WT & TC & ET & WT & TC & ET & WT & TC & ET \\
\hline
Fold 1 & 93.01 & 90.39 & 85.23 & \textbf{93.83} & 91.60 & 86.89 
& 93.57 \textcolor{red}{(-0.26)} & \textbf{92.68} \textcolor{green}{(+1.08)} & \textbf{87.87} \textcolor{green}{(+0.98)} \\
Fold 2 & 93.01 & 87.39 & 82.09 & \textbf{93.78} & 87.92 & 83.49 
& 93.14 \textcolor{red}{(-0.64)} & \textbf{88.91} \textcolor{green}{(+0.99)} & \textbf{84.08} \textcolor{green}{(+0.59)} \\
Fold 3 & 91.24 & 89.02 & 84.38 & \textbf{93.21} & \textbf{90.04} & 86.03 
& 92.36 \textcolor{red}{(-0.85)} & 89.90 \textcolor{red}{(-0.14)} & \textbf{86.32} \textcolor{green}{(+0.29)} \\
Fold 4 & 92.45 & 90.76 & 85.94 & \textbf{92.97} & 90.06 & 84.81 
& 92.63 \textcolor{red}{(-0.34)} & \textbf{91.76} \textcolor{green}{(+1.00)} & \textbf{87.76} \textcolor{green}{(+1.82)} \\
Fold 5 & \textbf{92.53} & 87.28 & 82.02 & 92.32 & 88.04 & 81.69 
& 92.56 \textcolor{green}{(+0.03)} & \textbf{90.19} \textcolor{green}{(+2.15)} & \textbf{85.29} \textcolor{green}{(+3.27)} \\
\hline
Mean & 92.45 & 88.97 & 83.93 & \textbf{93.22} & 89.53 & 84.58 
& 92.85 \textcolor{red}{(-0.37)} & \textbf{90.69} \textcolor{green}{(+1.16)} & \textbf{86.26} \textcolor{green}{(+1.68)} \\
Std & 0.72 & 1.62 & 1.80 & 0.62 & 1.55 & 2.06 
& \makebox[5em][l]{0.49} & \makebox[5em][l]{1.51} & \makebox[5em][l]{1.62} \\
\hline
\end{tabular}
\end{table*}
\begin{figure*}[t]
  \begin{subfigure}{0.48\textwidth}
\centerline{\includegraphics[width=\columnwidth]{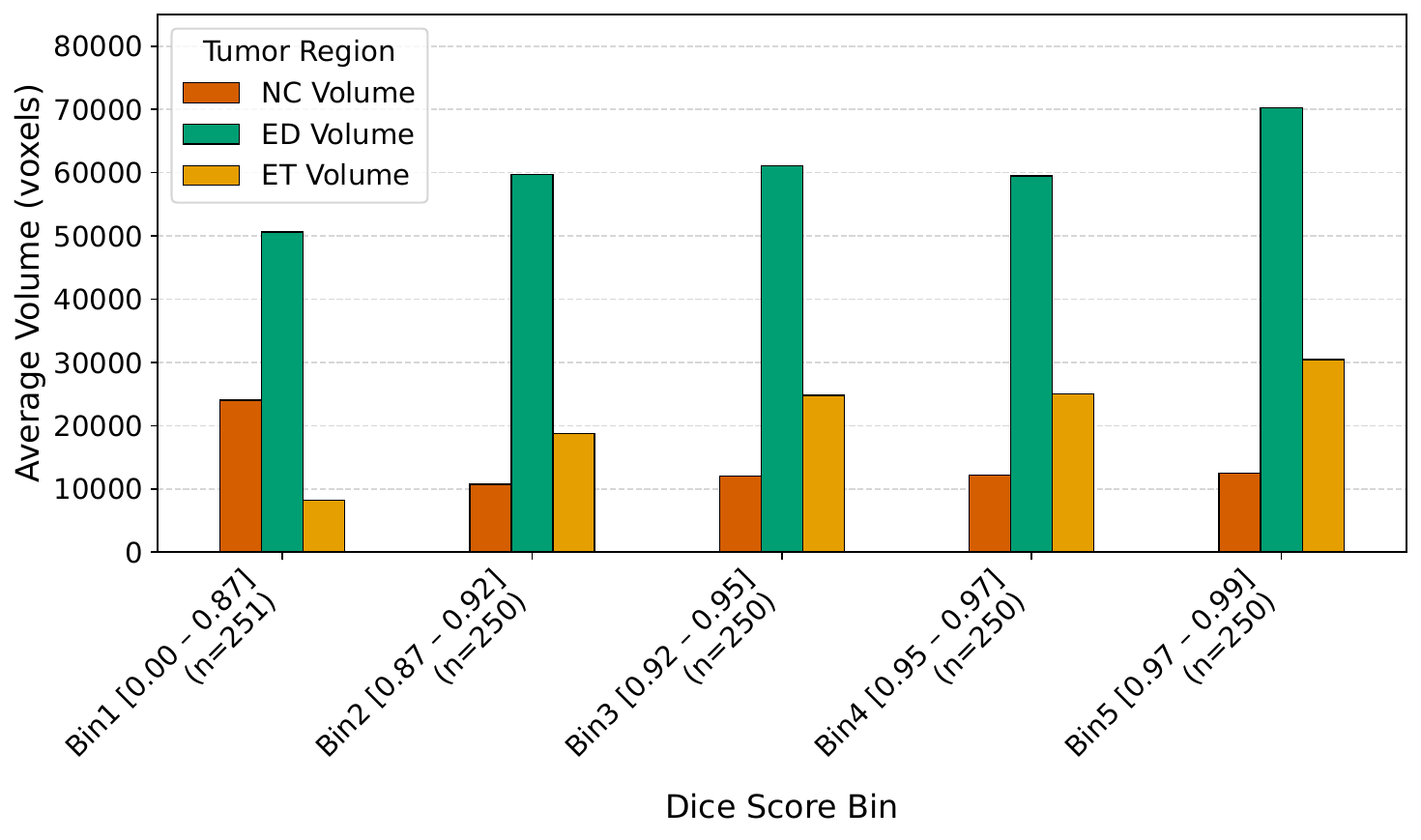}}
    \caption{}
    \label{fig:four-a}
  \end{subfigure}
  \begin{subfigure}{0.48\textwidth}
    \centerline{\includegraphics[width=\columnwidth]{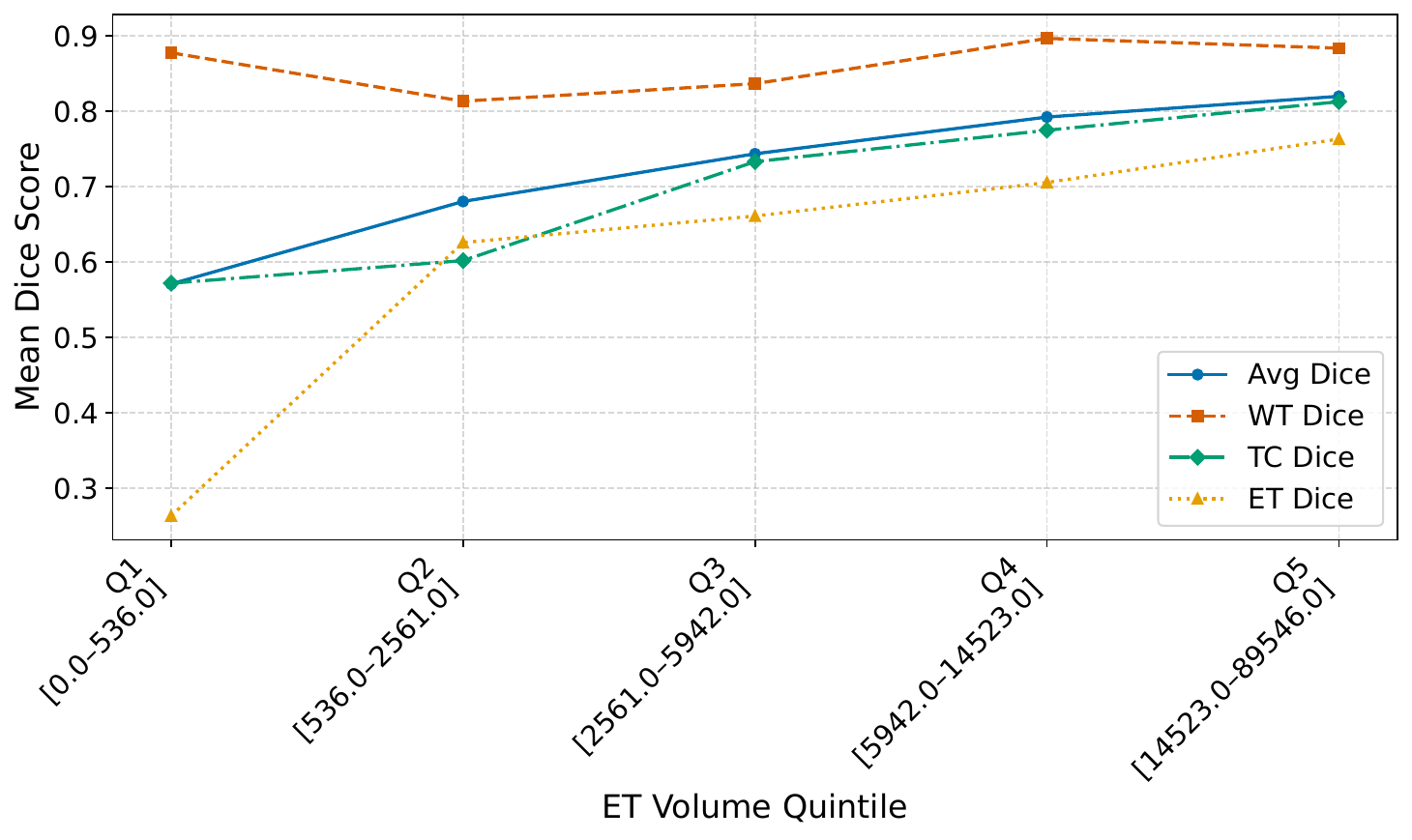}}
    \caption{}
    \label{fig:four-b}
  \end{subfigure}

  \caption{Comprehensive analysis of tumor heterogeneity and its impact on segmentation performance for the proposed systematic folds. 
  (a) Distribution of average tumor subregion volumes (ED, NC, ET) across Dice score bins,
  (b) Dice score variation across ET volume quintiles (Bin 1 only).}
  \label{fig:four}
\end{figure*}
In our proposed systematic five folds, all subjects are first grouped into five bins according to the average foreground intensity variation of tumor regions. From each bin, equal number of cases are sampled to form the five folds, ensuring balanced foreground intensity variation across the folds. While this strategy controls the foreground intensity variation, it naturally results in uneven representation of tumor subregion volumes (e.g., ET or NCR presence) across folds. 
Therefore, evaluation of different models on these systematic folds provides deeper insight into how tumor volume variations impact segmentation performance under balanced average foreground intensity variation across folds.

For each systematic fold, we train our proposed model and two state-of-the-art Mamba based segmentation models: SegMamba \cite{xing2024segmamba} and SegMambaV2 \cite{xing2025segmamba} for 300 epochs under identical hyperparameter settings, ensuring comparability across all the folds. These baselines are chosen as they achieve state-of-the-art performance with the single random 20\% test set as reported in Table \ref{tab:brats2023_results}, making them suitable references for testing their robustness in challenging heterogeneous systematic cross fold settings. 

Results are presented in Table \ref{tab:segmamba_comparison}. These results show that our model achieves superior Dice scores on the most challenging subregions (TC and ET), notably outperforming both SegMamba \cite{xing2024segmamba} and SegMambaV2 \cite{xing2025segmamba} across all proposed systematic five folds. On average, the proposed model achieves Dice scores of 92.85\% (WT), 90.69\% (TC), and 86.26\% (ET). The low standard deviations for all tumor subregions ($\leq$1.7\% Dice) across the systematic folds confirm the stability of the proposed model’s performance despite heterogeneous test distributions. WT segmentation is particularly robust (0.49\% Dice standard deviation), while TC and ET segmentation accuracy fluctuate more (standard deviations of 1.51\% and 1.62\% respectively), reflecting variation in tumor sizes across these systematic folds. 

To better understand the reason for the variation in results, we performed a detailed analysis linking the performance differences to each tumor subregion volume. We first grouped all test cases (1251 across folds) by their average Dice into five equal bins as presented in Fig. \ref{fig:four-a} (Bin1$\rightarrow$Bin5, having Dice score ranges mentioned on the x-axis) to examine how performance relates to tumor volume. It can be observed that ET volume increases steadily in bins as the Dice score increases, while both NC and ED show no specific trend with the increase of Dice score. These observations clearly indicate that ET volume is the most dominant factor influencing segmentation performance. 

To further examine the relationship of Dice score and tumor volume more closely, we subdivided the lowest performing group (Bin1) into five quintiles (Q1$\rightarrow$Q5) according to ET volume. It can be observed from Fig. \ref{fig:four-b} that Dice scores improve progressively from Q1 to Q5 across tumor subregions (TC, ET, and WT). The small ET volume cases pose the greatest challenge, whereas those with larger ET volumes are segmented more accurately. 

This detailed analysis of our model’s performance, along with the comparison between random and systematic folds, highlights that the proposed systematic folds establish a structured protocol for evaluating model robustness across varying difficulty levels, from easier (Folds 1 and 4) to intermediate (Folds 3 and 5) and more difficult test distributions (Fold 2). Moreover, these systematic folds provide clear insight that the variation in segmentation performance for clinically relevant tumor sub-regions (TC and ET) across these folds arises primarily from differences in enhancing tumor volumes, as intensity variation is already controlled across folds. 
\begin{figure}[!t]
\centerline{\includegraphics[width=\columnwidth]{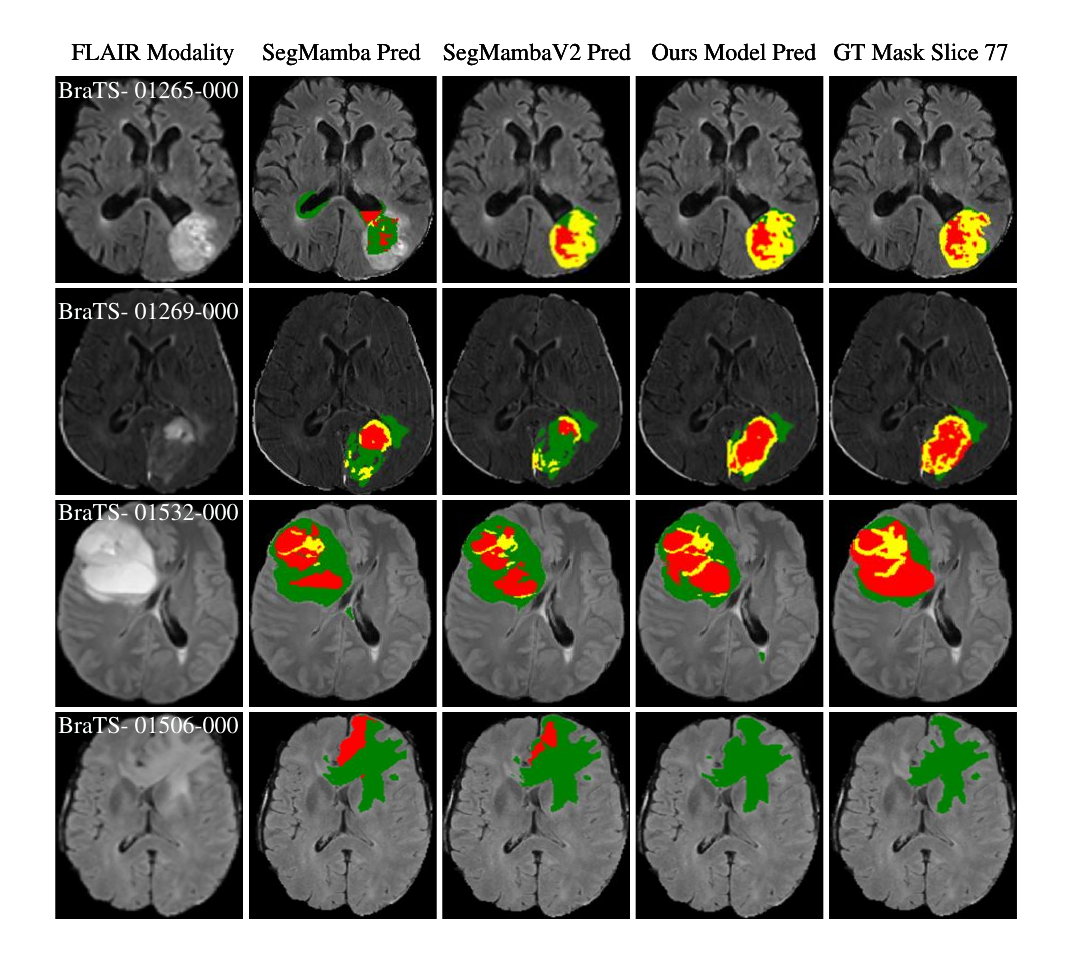}}
\caption{Qualitative performance comparison of SegMamba, SegMambaV2, and the proposed method. Mid-slices of BraTS test cases are shown, with ground-truth masks and models' predictions. Tumor subregions are indicated as edema (green), necrotic core (red), and enhancing tumor (yellow).}
\label{fig:qualitative}
\end{figure}
\begin{figure}[]
\centerline{\includegraphics[width=\columnwidth]{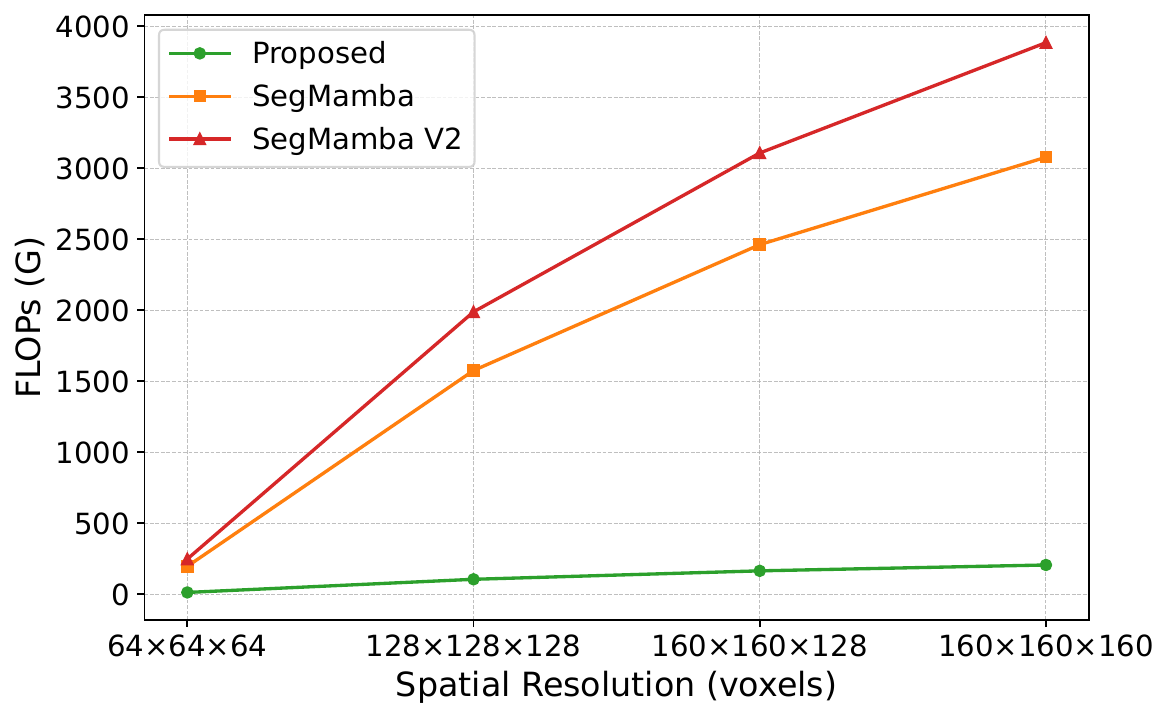}}
\caption{Comparison of computational complexity of the proposed model and state-of-the-art models with respect to the increase in spatial resolution in BraTS2023.}
\label{comcost}
\end{figure}

In addition to the quantitative analysis, qualitative results offer further insights into segmentation performance. Fig. \ref{fig:qualitative} illustrates middle slices from systematic Fold 3 test cases, comparing predictions from SegMamba \cite{xing2024segmamba}, SegMambaV2 \cite{xing2025segmamba}, and the proposed model. The visualizations show that our proposed model achieves more precise delineations of the enhancing tumor (yellow) and necrotic core (red), whereas the other models under-segment or miss small regions. The edema boundaries (green region) are also well preserved, highlighting the model’s ability to maintain accuracy across all tumor sub regions even under difficult conditions. 

Complementing these findings, we compared computational efficiency of the models in Fig. \ref{comcost}. We can see that FLOPs of SegMamba \cite{xing2024segmamba} and SegMambaV2 \cite{xing2025segmamba} increase drastically with increasing spatial resolution, whereas our model maintains a significantly more efficient computational scaling. This demonstrates the computational efficiency of our method as compared to other methods, while achieving strong accuracy, particularly at higher resolutions.

To further quantify the efficiency advantage, Table \ref{tab:comp_comp} summarizes detailed model statistics and runtime performance. The proposed model follows a lightweight design, containing only 29M parameters, compared to 64M and 136M parameters for SegMamba \cite{xing2024segmamba} and SegMambaV2 \cite{xing2025segmamba}, respectively.
Runtime performance is evaluated on identical hardware using a single NVIDIA RTX 3090 GPU, where the inference time of the proposed model is 2.50s per sample, while SegMamba and SegMambaV2 require 9.13s and 11.11s, respectively. In addition, at a spatial resolution of $128^3$, the proposed model incurs only 105G FLOPs, whereas SegMamba \cite{xing2024segmamba} and SegMambaV2 \cite{xing2025segmamba} require 1575G and 1985G FLOPs respectively, corresponding to approximately $15\times$ and $19\times$ lower computational
cost for the proposed model relative to SegMamba and SegMambaV2.

\begin{table}[]
\centering
\caption{Comparison of model complexity and runtime performance between the proposed model and state-of-the-art Mamba-based segmentation methods.}
\label{tab:comp_comp}
\setlength{\tabcolsep}{3pt}
\begin{tabular}{l|c|ccc}
\hline
Methods & Params (M) & FLOPs (G) 
& Infrence Time (s) \\
\hline
SegMamba \cite{xing2024segmamba} & 64 & 1575 & 9.13 \\
SegMambaV2 \cite{xing2025segmamba} & 136 & 1985 & 11.11 \\
Proposed & \textbf{29} & \textbf{105} & \textbf{2.50} \\
\hline
\end{tabular}
\end{table}

\subsection{Ablation Study}
We conducted an ablation study on a randomly selected fold (Fold 2) of BraTS2023 to evaluate the impact of each architectural component of our proposed method.
\subsubsection{Effect of Space-Filling Curve (SFC) and Gated Fusion (GF)}
The effect of Mamba directional modeling, space-filling curve (SFC) mapping, and gated
fusion (GF) is evaluated on systematic Fold2 of BraTS2023, with quantitative results
reported in Table~\ref{tab:ablation_fold2}. The uni-directional Mamba baseline, without gated fusion, provides a strong starting point (WT: 93.07, TC: 87.80, ET: 83.13). Extending to a bi-directional setup while fusing forward and reverse features through simple summation (SFC $\checkmark$, GF $\times$) yields marginal gains in TC and ET accuracy, suggesting that naive feature fusion does not fully exploit bi-directional information. Conversely, enabling gated fusion (GF $\checkmark$) while omitting SFC ($\times$) improves WT accuracy but degrades TC and ET, highlighting the importance of SFC in preserving positional consistency and stabilizing feature interactions. The full design, combining bi-directional Mamba with both SFC and gated fusion (SFC $\checkmark$, GF $\checkmark$), delivers the best overall performance (WT: 93.14, TC: 88.91, ET: 84.08) and lower HD95 across tumor subregions.
\begin{table}[]
\centering
\caption{Ablation study on the proposed systematic Fold2 of BraTS2023 evaluating the
effects of space-filling curve (SFC) mapping and gated fusion (GF) in uni and
bi-directional Mamba configurations. Dice (\%) and HD95 (mm) are reported for
whole tumor (WT), tumor core (TC), and enhancing tumor (ET).}
\label{tab:ablation_fold2}
\setlength{\tabcolsep}{3pt}
\begin{tabular}{c|c|c|cccccc}
\hline
Mamba & \multirow{2}{*}{SFC} & \multirow{2}{*}{GF} 
& \multicolumn{2}{c}{WT} & \multicolumn{2}{c}{TC} & \multicolumn{2}{c}{ET} \\
\cline{4-9}
Configuration & & & Dice & HD95 & Dice & HD95 & Dice & HD95 \\
\hline
Uni-directional & $\checkmark$ & $\times$ & 93.07 & 5.88 & 87.80 & 5.69 & 83.13 & 6.25 \\
Bi-directional & $\checkmark$ & $\times$ & 92.97 & 6.71 & 88.58 & 6.33 & 83.57 & 6.50 \\
Bi-directional & $\times$ & $\checkmark$ & 93.26 & 6.96 & 87.34 & 6.76 & 82.94 & 7.11 \\
Bi-directional & $\checkmark$ & $\checkmark$ & 93.14 & 6.31 & 88.91 & 5.63 & 84.08 & 6.06 \\
\hline
\end{tabular}
\end{table}

\subsubsection{Effect of Vector Quantization on Robustness}
The effect of the vector quantization (VQ) block on segmentation robustness is
evaluated under both clean and noise-corrupted test conditions, with results
summarized in Table~\ref{tab:ablation_fold2_vq}. Under clean evaluation
conditions (100\% signal), where the signal corresponds to the original
multi-modal MRI inputs, the inclusion of vector quantization (VQ $\checkmark$) leads to consistent performance improvements across all tumor subregions, resulting in Dice gains of +0.30\%, +1.38\%, and +0.50\% for WT, TC, and ET, respectively, compared to the non-quantized configuration (VQ $\times$).

When the input MRI modalities are corrupted with Gaussian noise at equal
signal-to-noise ratios (50\% signal / 50\% noise), the non-quantized model (VQ $\times$) exhibits substantial performance degradation, indicating high sensitivity to input perturbations. In contrast, the VQ-enabled model (VQ $\checkmark$) significantly mitigates this degradation relative to its non-quantized counterpart (VQ $\times$). Specifically, under noisy conditions, the quantized model (VQ $\checkmark$) maintains Dice scores that are higher by 9.49\%, 3.10\%, and 3.02\% for WT, TC, and ET, respectively, compared to the non-quantized configuration (VQ $\times$), while also yielding consistently lower HD95 across all tumor regions.

Overall, this ablation demonstrates the importance of the VQ block in stabilizing latent representations under noisy input conditions. By constraining features to a discrete codebook, vector quantization reduces sensitivity to noise while preserving discriminative information, leading to improved segmentation robustness.
\begin{table}[]
\centering
\caption{Ablation study on the proposed systematic Fold2 of BraTS2023 analyzing the
impact of the vector quantization (VQ) block under clean and noisy test conditions.
Dice (\%) and HD95 (mm) are reported for whole tumor (WT), tumor core (TC), and
enhancing tumor (ET).}
\label{tab:ablation_fold2_vq}
\setlength{\tabcolsep}{2pt}
\begin{tabular}{c|c|c|c|cccccc}
\hline
Mamba & \multirow{2}{*}{VQ} & \multirow{2}{*}{Signal} & \multirow{2}{*}{Noise} 
& \multicolumn{2}{c}{WT} & \multicolumn{2}{c}{TC} & \multicolumn{2}{c}{ET} \\
\cline{5-10}
Configuration & & & & Dice & HD95 & Dice & HD95 & Dice & HD95 \\
\hline
Bi-directional & $\times$ & 100\% & 0\% & 92.84 & 5.26 & 87.53 & 6.15 & 83.58 & 6.47 \\
Bi-directional & $\checkmark$ & 100\% & 0\% & 93.14 & 6.31 & 88.91 & 5.63 & 84.08 & 6.06 \\
Bi-directional & $\times$ & 50\% & 50\% & 76.79 & 13.05 & 71.83 & 13.83 & 70.71 & 13.57 \\
Bi-directional & $\checkmark$ & 50\% & 50\% & 86.28 & 9.39 & 74.93 & 12.92 & 73.73 & 12.04 \\

\hline
\end{tabular}
\end{table}
\section{Conclusion} 
This paper proposed a novel dual-resolution bi-directional Mamba architecture for efficient and reliable brain tumor segmentation. The proposed model leverages space-filling curve, gated fusion, and vector quantization to capture multi-scale sequence dependencies. Experimental results validate that our model outperforms state-of-the-art baselines models across all tumor subregions while maintaining minimal computational overhead, highlighting its clinical applicability. Moreover, we proposed  systematic five-fold partitioning of the BraTS2023 dataset to ensure fair and comprehensive assessment. This enables rigorous evaluation across diverse data distributions, while explicitly accounting for scanner-dependent intensity variations. Our detailed analysis across these systematic folds reveals that imprecise segmentation predominantly occur in cases with very small enhancing tumor volumes. The performance comparison demonstrates that our proposed model consistently achieves superior performance over state-of-the-art baselines on the proposed systematic folds, marking a step towards more reliable tumor segmentation in heterogeneous clinical settings.

\bibliographystyle{IEEEtran}
\bibliography{references}

\end{document}